\documentclass[lettersize,journal]{IEEEtran}
\usepackage{amsmath,amsfonts}
\usepackage{algorithmic}
\usepackage{algorithm}
\usepackage{threeparttable}
\usepackage{array}
\usepackage[caption=false,font=normalsize,labelfont=sf,textfont=sf]{subfig}
\usepackage{textcomp}
\usepackage{tabularx}
\usepackage{multirow}
\usepackage{booktabs}
\usepackage{stfloats}
\usepackage{url}
\usepackage{orcidlink}
\usepackage{verbatim}
\usepackage{graphicx}
\usepackage{cite}
\usepackage{enumerate}
\usepackage{hyperref}
\hypersetup{hidelinks,linkcolor=subsectioncolor}
\begin{document}

\title{Faces of the Mind: Unveiling Mental Health States Through Facial Expressions in 11,427 Adolescents}

\author{Xiao Xu\orcidlink{0009-0004-0917-5867}, Xizhe Zhang\orcidlink{0000-0002-8684-4591}, Yan Zhang

\thanks{This study was funded by the National Natural Science Foundation of China (62176129), the National Key Research and Development Program (2022YFC2405603). \textit{(Corresponding authors: Xizhe Zhang: \href{mailto:zhangxizhe@njmu.edu.cn}{zhangxizhe@njmu.edu.cn})}}
\thanks{Xizhe Zhang is with the Early Intervention Unit, Department of Psychiatry, The Affiliated Brain Hospital of Nanjing Medical University, and the School of Biomedical Engineering and Informatics of Nanjing Medical University, Nanjing, P.R. China.}
\thanks{Xiao Xu, Yan Zhang are with the School of Biomedical Engineering and Informatics, Nanjing Medical University, Nanjing, P.R. China. (e-mail: \href{mailto:xuxiaooo1111@gmail.com}{xuxiaooo1111@gmail.com})}
\thanks{All facial images used in this paper were generated by Stable Diffusion \cite{rombach2021highresolution} and do not represent real individuals.}}

\markboth{Preprint Submitted to IEEE Transactions on Affective Computing}%
{Shell \MakeLowercase{\textit{Xiao Xu et al.}}: FACES of the Mind: Unveiling Mental Health States Through Facial Expressions in 11,427 Adolescents}

\maketitle

\begin{abstract}
Mood disorders such as depression and anxiety often manifest through facial expressions, but existing machine learning algorithms designed to assess these disorders have been hindered by small datasets and limited real-world applicability. To address this gap, we analyzed facial videos of 11,427 participants - a dataset two orders of magnitude larger than those used in previous studies - including standardized facial expression videos and psychological assessments of depression, anxiety, and stress. However, scaling up the dataset introduces significant challenges due to increased symptom heterogeneity, making it difficult for models to learn accurate representations. To address this, we introduced the Symptom Discrepancy Index (SDI), a novel metric for quantifying dataset heterogeneity caused by variability in individual symptoms among samples with identical total scores. By removing the 10\% most heterogeneous cases as identified by the SDI, we raised the F1 scores of all models from approximately 50\% to 80\%. Notably, comparable performance gains were observed within both the retained and excluded subsets. These findings demonstrate symptom heterogeneity, not model capacity, as the principal bottleneck in large scale automated assessment and provide a general solution that is readily applicable to other psychometric data sets.

\end{abstract}


\section{Introduction}

\IEEEPARstart{T}{he} global burden of common mental disorders has increased steadily in the past two decades, and the COVID-19 pandemic produced an abrupt inflection in this trajectory: Recent meta-analyses estimate prevalence jumps of roughly 24.4\% and 22.9\% for depression and anxiety in the general population, with even larger surges among children and adolescents \cite{nochaiwong2021global,bower2023hidden}. Early detection and continuous monitoring are therefore critical public health imperatives, yet the field still relies on lengthy clinician‑administered interviews such as the Structured Clinical Interview for DSM (SCID) \cite{glasofer2015structured} or on observer‑rated scales such as the Hamilton Depression Rating Scale (HAMD) \cite{bagby2004hamilton}. These instruments consume scarce specialist time and are inevitably colored by rater subjectivity. An objective, scalable complement capable of operating in nonclinical environments remains an open challenge.

Facial expressions constitute a promising behavioral channel for such an objective assessment. They convey fine‑grained information about affective states and have long been examined in psychiatric examinations \cite{chen_emotional_2000,elfenbein_emotional_2002}. Recent advances in computer vision have produced deep learning models that learn spatiotemporal facial patterns indicative of major depressive disorder, using architectures that span static 2D-CNNs \cite{de2019depression}, sequential encoders such as LSTMs \cite{uddin2020depression}, and multiscale 3D-CNNs with attention mechanisms \cite{he2022intelligent}. However, most published studies are trained and validated on small corpora such as  DAIC-WOZ \cite{gratch_distress_2014} and AVEC \cite{valstar2013avec, valstar_avec_2014} , each containing a few hundred individuals. Reported accuracies sometimes exceed 90\% \cite{yin2023depression,gupta2024depression}, but these figures rarely translate into real-world settings, indicating overfitting to narrow population strata and recording conditions.

A second, less recognized obstacle concerns the way ground‑truth labels are constructed. Self‑report instruments such as the PHQ \cite{kroenke2001phq} or DASS‑21 \cite{lovibond1995depression} convert a heterogeneous collection of symptoms into a single total score that is then divided into two or more severity categories. However, mental disorders are highly heterogeneous. Depression, for instance, includes various subtypes and frequently co-occurs with other conditions like anxiety \cite{fried2014impact}. Patients with identical sums may endorse entirely different item subsets, a phenomenon that introduces systematic label noise. Standard supervised learning mixes up these different profiles, inflating intra‑class variance and distorting decision boundaries.

To address these challenges, we present FACES (Facial Analysis Collection for Emotional States), a comprehensive dataset comprising high-quality facial videos specifically tailored for assessing mental health in adolescents. The initial phase of FACES has already collected data from 11,427 participants, with plans to expand to over 100,000 individuals in subsequent phases. FACES is part of the SEARCH cohort study \cite{Zhange300861}, which collects facial videos, audio recordings, and diverse scales. This large-scale longitudinal study involves students, caregivers, and teachers across 11 schools in Jiangsu, China, focusing on emotional well-being, sleep patterns, risk behaviors, family environment, trauma exposure, and academic performance. To our knowledge, FACES represents the largest and most standardized facial dataset for adolescent mental health. 

Leveraging the FACES dataset, we evaluated the performance of a spectrum of methods, spanning conventional machine-learning algorithms and state-of-the-art deep learning algorithms. Surprisingly, all models converged on similar accuracies of approximately 50–60\%, implying that they failed to learn clinically meaningful patterns. We therefore proposed the Symptom Discrepancy Index (SDI), which measures the degree of symptom variance among instances that share an identical label. Incorporating SDI into the modeling pipeline, by preferentially retaining observations with low discrepancy, led to a striking improvement, raising the F1 score from roughly 0.5 to 0.8. This result indicates that symptom-driven heterogeneity, rather than limited model capacity, underlies the disappointing outcomes observed on large-scale affective-computing corpora and explains why models that perform well on smaller, more homogeneous samples falter when applied to broader populations.

Considering the discussion above, the contributions of this research are:
\begin{enumerate}
\item \textbf{Large Benchmark Dataset:} We introduce FACES, a corpus that comprises 11,427 pupils aged ten to eighteen who were recorded with identical tablet hardware while reading a neutral passage under controlled illumination. After stringent temporal, visual and behavioral quality screens, 8,281 five‑minute videos remain, each paired with a time‑stamped DASS‑21 questionnaire. The scale, demographic breadth and methodological uniformity of FACES provide unprecedented statistical power for training deep architectures, probing domain‑shift robustness, and conducting longitudinal follow‑ups as subsequent waves of data are added.

\item \textbf{Novel Discrepancy Index:} We introduced a novel metric, the Symptom Discrepancy Index (SDI), a theoretically grounded statistic that captures the spread of item‑level response patterns among participants who obtain the same summed score. Unlike existing ad‑hoc heuristics, SDI is scale‑invariant, easily computed, and exposes the latent heterogeneity that arises when multifaceted symptom constellations are collapsed into coarse severity bands.

\item \textbf{Impact of Label Aggregation:} We identified the use of total scores from psychometric scales as a key factor contributing to diminished model performance. Total scores aggregate individual item responses, masking the heterogeneity among scale items and preventing models from capturing the underlying constructs. Through the selective removal of a small number of highly biased samples, we demonstrated a marked improvement in the performance of all machine learning models. This finding underscores the importance of addressing label heterogeneity to enhance the accuracy and generalizability of machine learning models.
\end{enumerate}

The remainder of this article proceeds as follows. Section II surveys related work on facial analysis for mental‑health assessment and on label heterogeneity in psychometrics. Section III details the FACES corpus, the SDI formulation, and the experiment pipeline. Section IV presents classification and regression experiments that quantify the impact of SDI‑based filtering. Finally, Section V discusses concluding remarks and outlines future research directions.

\section{Literature Survey}

\subsection{Label Noise from Scale Heterogeneity}
As machine learning models increasingly rely on large-scale clinical and research datasets, the heterogeneity of psychiatric rating scales has emerged as a critical bottleneck for reliable prediction. In a review of 450 randomized controlled trials for depression, 388 distinct outcome measures were used, with only 40\% employing the classic HAMD—highlighting the lack of standardized labeling across datasets and hindering models from learning stable depressive representations \cite{veal2024heterogeneity}. Content analyses further reveal that within the same diagnosis, symptom overlap across questionnaires can be as low as 29\%, while 60\% of symptoms recur across different disorders. This undermines label consistency within a single category and introduces spurious comorbidity noise from the model’s perspective \cite{newson2020heterogeneity}. Scale-specific structural instability exacerbates the issue: PHQ-9 has been modeled with one-, two-, and even four-factor solutions depending on the sample, meaning the same total score may represent different latent traits across datasets \cite{lamela2020systematic}. Similarly, a COSMIN-based meta-analysis of 119 studies found the canonical three-subscale PANSS model to lack structural validity despite acceptable reliability, suggesting that schizophrenia-related labels may be systematically biased \cite{geck2025cosmin}. Even widely used instruments like the PHQ-9 and GAD-7 have only recently demonstrated strict sex invariance in a large clinical sample, while cross-national studies show that the BAI fails measurement invariance across Spanish, Portuguese, and Brazilian populations—raising the risk that models might misinterpret cultural differences as pathological features \cite{saunders2023measurement, do2023psychometric}. Taken together, heterogeneity in scale selection, factor structure, and population validity contributes to label noise, target shift, and out-of-distribution generalization failure. This underscores the urgent need to construct cross-scale mappings, apply measurement invariance corrections, and adopt psychometrically standardized core outcome sets before deploying machine learning in psychiatric prediction.

\subsection{Facial Expressions as Mental Disorder Biomarkers}

Facial expressions have emerged as critical behavioral markers in psychiatric research, reflecting the intrinsic link between affective display and mental states. Davies et al. \cite{davies_facial_2016} reviewed consistent alterations in emotional expressions across non-psychotic disorders, suggesting shared affective disruptions. Extending this, Namba et al. \cite{namba_spontaneous_2017} identified novel sadness-related facial expressions beyond traditional emotional categories, pointing to the need for more nuanced interpretation. Gupta et al. \cite{gupta_alterations_2019} further demonstrated that facial expression changes in early youth may signal prodromal psychosis, highlighting their predictive value. Parallel efforts in oculomotor research have revealed complementary biomarkers: schizophrenia has been linked to abnormal fixational saccades reflecting cognitive and positive symptoms \cite{liu2024spatial,okazaki2023discrimination}, while gaze-related features have shown diagnostic potential for depression \cite{gao2023abnormal}. These findings collectively support the utility of facial and eye movement cues as accessible, quantifiable indicators for early screening and clinical evaluation.

\subsection{Mental Disorder Detection Based on Facial Expressions}

Automated analysis of facial behavior in mental health has progressed through distinct methodological stages. Early work relied on traditional machine learning models—such as support vector regression \cite{meng2013depression}, decision trees \cite{nasir2016multimodal}, and logistic regression \cite{alghowinem2013head}—in combination with hand-crafted features like LBP \cite{jan2017artificial, zhu_automated_2017}, LLD \cite{nasir2016multimodal}, and HOG \cite{valstar_avec_2014}. These approaches, while foundational, were limited in capturing temporal and multimodal dynamics.

To address this, subsequent studies adopted probabilistic modeling techniques such as Gaussian Mixture Models (GMMs). Williamson et al. \cite{williamson2014vocal} modeled vocal tract dynamics using formants and delta-mel-cepstra, while others applied GMM-UBM for multimodal fusion \cite{cummins2013diagnosis} or employed Fisher Vector encoding from segmented videos \cite{jain2014depression}.

Recent advances center on deep learning frameworks. CNNs and RNNs have enabled more expressive and temporally-aware representations \cite{yang_attention_2023, xu_attention-based_2023}. Al Jazaery et al. \cite{al2018video} used C3D networks with RNNs to model short-term and sequential dynamics. Song et al. \cite{song_spectral_2020} introduced spectral representations of behavioral primitives, enabling multi-scale temporal modeling. Zhu et al. \cite{zhu_automated_2017} and Wen et al. \cite{wen2015automated} applied deep convolutional networks and sparse coding for dynamic facial analysis. A more recent multimodal framework by He et al. \cite{he_deep_2022} fused 3D-CNN-based facial features and BiLSTM-modeled audio, achieving state-of-the-art results through deep feature integration.

Despite promising performance, challenges remain in clinical translation. Current models often lack robustness across contexts, suffer from limited interpretability, and rely on constrained data environments. Future research should emphasize real-world validation, explore vision-language integration, and leverage advances in multimodal large-scale pretraining to enhance model generalization and applicability in healthcare.

\subsection{Psychiatric-related Facial Datasets}
A number of datasets have been established for the study of emotions and mental disorders in recent years, reflecting the growing interest and advancements in this domain. These datasets encompass a wide range of subjects and modalities, which allow researchers to explore multifaceted dimensions of emotion and clinical states.

\textbf{AVEC2013} \cite{valstar_avec_2013} \& \textbf{AVEC2014} \cite{valstar_avec_2014}: The two corpora are key benchmarks for depression analysis. The AVEC2013 dataset comprises 340 videos of human-computer interactions from 292 participants. Its successor, AVEC2014, extended this corpus by introducing two additional tasks, ``Freeform" and ``Northwind," resulting in a combined total of 300 available data samples across the challenges. Both datasets consistently employ the Beck Depression Inventory-II (BDI-II) scale for labeling depression severity.

\textbf{BD} \cite{cciftcci2018turkish}: Comprising 46 patients and 49 healthy controls, the data were collected from the mental health service of a hospital. The participants underwent semi-structured interviews, and their depressive and manic features were evaluated and recorded over specific intervals. This database was later employed as challenge data in AVEC2018.

\textbf{DAIC-WOZ} \cite{gratch_distress_2014}: This U.S.-based database features 189 sessions of interactions captured in multiple modes like Face-to-Face, Teleconference, and Wizard-of-Oz, among others. It not only comprises audiovisual cues but also includes physiological data like GSR, ECG, and respiration. It was used as the AVEC2017 dataset. The DAIC has been further expanded with an extended version known as the DAIC-WOZ-extend, and employed as challenge data in AVEC2019, currently being one of the most popular open-access databases in the domain.

\textbf{Pittsburgh} \cite{dibekliouglu2017dynamic}: This dataset consists of data from 57 participants (34 females and 23 males) undergoing clinical treatment for depression. The subjects, aged between 19 and 65 years, were assessed over a duration of 21 weeks using clinical interviews. The data includes facial video recordings from these interviews, providing a rich resource for analyzing expressions and behaviors related to depression. The severity of symptoms was evaluated using the Hamilton Rating Scale for Depression (HRSD).

\textbf{eFEV-PW}\cite{li2024automatic}: This dataset includes 65 patients with depression. Emotional facial expression videos (eFEVs) and pupil-wave signals (ePW) were collected while participants watched emotionally stimulating videos. Combined with HAMD and PHQ-9 scores, deep learning models (3DCNN+LSTM and 1DCNN) were used for depression assessment.

\textbf{Facial Cues from Emotional Stimuli}\cite{hu2023detecting}: This dataset comprises 62 participants (31 patients with depression and 31 healthy controls). Emotional stimuli from film clips were used to collect facial expression units (AUs) and head movement features. PHQ-9 and PHQ-15 scores were utilized to construct a depression detection framework based on local facial feature reconstruction.

\textbf{MuSE}\cite{jaiswal2020muse}: This dataset involves 28 college students from the University of Michigan, with recordings collected under high-stress (during final exams) and low-stress (post-finals) conditions. Multimodal features, including audio, video, thermal imaging, and physiological signals, were recorded. PSS and SAM scales were used to study the interplay between stress and emotions.

These datasets, offering diverse modalities and demographics, are invaluable for researching emotional states and mental disorders. Some of these datasets serve as widely used public datasets, providing a research foundation for the introduction of new deep learning methods. However, they also face limitations due to small sample sizes and the reliance on single assessment scales. To better support new methodologies and facilitate the clinical application of effective methods, future research will require larger datasets and more comprehensive assessment tools.

\section{Materials And Methods}

\begin{figure*}[htbp]
\centering
\includegraphics[width=1.6\columnwidth]{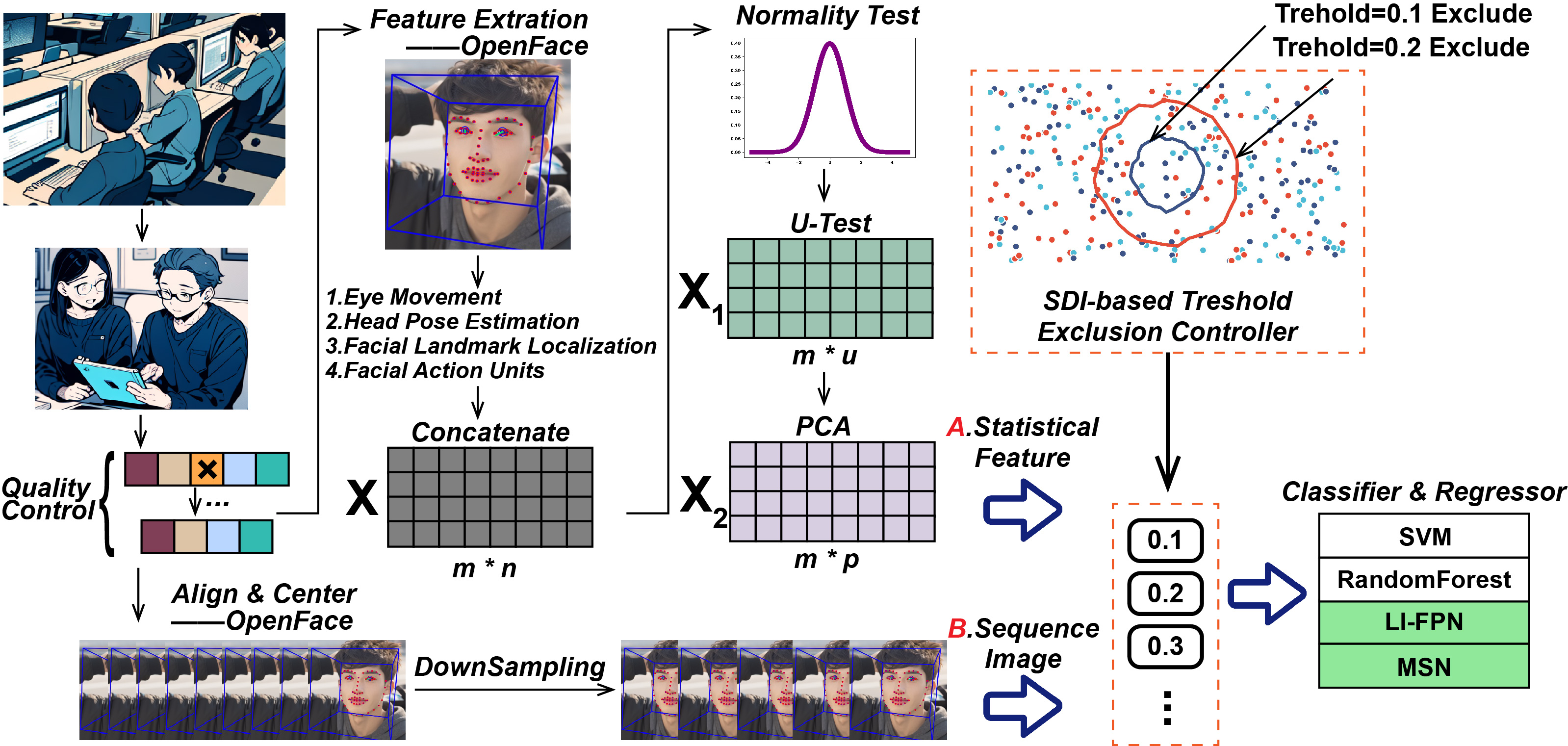}
\caption{\textbf{Main Workflow.} In the data collection phase, we collected questionnaire and video data, followed by stringent quality control procedures. We began by extracting facial features using the OpenFace tool, calculating their statistical values, and concatenating them to create a comprehensive dataset, denoted as $X$. We then conducted regression analyses on $X$. During the data preprocessing stage, we first determined the applicability of the U-test based on a normality assessment of $X_1$. This was followed by dimensionality reduction through Principal Component Analysis (PCA) to form $X_2$, which is our Statistical Feature. Next, we performed face alignment centered on the video data, downsampled the sequential image frames on a frame-by-frame basis to form the Sequential Image feature. Finally, we use traditional machine learning models (for Statistical Features) and SOTA deep learning models (for Sequence Images) with our SDI-based sample filter for classification and regression evaluation.}
\label{fig1}
\end{figure*}

This section details the dataset, the Symptom Discrepancy Index (SDI) that underpins all later analyses, and the full processing pipeline from raw video to predictive modeling. A schematic overview is provided in \textbf{Figure \ref{fig1}}.

\subsection{The FACES Dataset}
\subsubsection{Study Design and Recruitment} The dataset is part of the School-based Evaluation Advancing Response for Child Health (SEARCH) \cite{Zhange300861}, a comprehensive, mixed-method longitudinal cohort study designed to meet the growing needs of individuals seeking access to mental health care services. Our study focuses on the mental and emotional well-being of primary and secondary school students aged 10 to 18, spanning grades 4 to 12. This collaboration serves to foster a comprehensive approach to studying children and adolescent mental well-being within a school-based framework. Ultimately, 11,427 students were successfully recruited between September 28, 2022, and November 1, 2022.

\subsubsection{Data Collection Methodology} Data collection was carefully made consistent to maintain uniformity. Specially trained investigators managed and oversaw the digital platform designed for this task. At first, students were guided to fill out scales on a dedicated website in the school's computer classrooms. Later on, students moved to a quieter setting, where local investigators helped gather facial and audio samples using a specific app on Android tablets.

The main psychological state assessment tool for FACES was the DASS-21. This simplified version of the Depression Anxiety Stress Scales is segmented into three categories, Depression, Anxiety, and Stress, each consisting of seven specific items. The response to each item is measured on a four-point Likert scale, which facilitates the categorization of the severity of the symptoms in the mental health spectrum of interest. Sub‑scale totals were mapped to the five conventional severity bands: Normal, Mild, Moderate, Severe and Extremely‑Severe. The cut‑offs follow the manual guidelines: 10, 14, 21 and 28 for depression; 8, 10, 15 and 20 for anxiety; 15, 19, 26 and 34 for stress.

\subsubsection{Video Recording Protocol} After completing the questionnaire in a computer laboratory, each pupil moved to a quiet room with controlled illumination. Video recording employed a Lenovo Xiaoxin Pad TB‑J606 tablet running bespoke software that guided pupils to read the neutral story “The North Wind and the Sun”. Cameras were positioned approximately 20cm from the face; sessions lasted five minutes and were stored as H.264 streams at 1280×720 pixels and 30 fps.

\subsubsection{Data Quality Control} Four exclusion criteria were applied sequentially. (i) Recordings shorter than 20 seconds or containing more than five seconds of dropped frames were discarded. (ii) OpenFace2.0 failed detections triggered removal if no complete face was present for more than ten consecutive frames, if the face was occluded (e.g. mask) or located outside the frame. (iii) Response‑time validity checks removed questionnaire entries whose mean item latency was below 0.5s or above 20s \cite{su2024exploring, su2024temporal}. (iv) A final manual review confirmed that questionnaire and video belonged to the same pupil and that lighting conditions were adequate.

\subsubsection{Population Distribution} Application of the above filters yielded a final analytic sample of 8,281 pupils. The mean age was 13.47 years (SD=2.35); 52.7\% identified as male. Grade, age and gender counts are reported in \textbf{Table} \textbf{\ref{table_population_distribution}}. \textbf{Table} \textbf{\ref{table_DASS_21}} summarizes the DASS‑21 severity distribution: for depression, 6316 pupils fell in the Normal band and 207 in the Extremely‑Severe band; analogous right‑skewed patterns were observed for anxiety and stress. This imbalance reflects the prevalence of sub‑clinical presentations in community samples and motivated the heterogeneity analyses.

\begin{table}[htbp]
\caption{Grade Distribution.}
\label{table_population_distribution}
\centering
\setlength{\tabcolsep}{1mm}{
\scriptsize
\begin{tabular}{c c c}
\toprule
\textbf{Characteristics} & \textbf{No. of Samples} & \textbf{Percentage}\\
\midrule
\textit{Gender} & & \\
\midrule
male & 4367 & 52.74 \\
female & 3914 & 47.26 \\
\midrule
\textit{Age} & & \\
\midrule
Mean age (SD) & 13.47 & 2.35 \\
10 & 1157 & 13.97  \\
11 & 957 & 11.56  \\
12 & 1119 & 13.51  \\
13 & 958 & 11.57  \\
14 & 992 & 12.00  \\
15 & 1031 & 12.45  \\
16 & 1050 & 12.68  \\
17 & 875 & 10.57  \\
18 & 142 & 1.71  \\
\midrule
\textit{Grade} & & \\
\midrule
4 & 200 & 2.42  \\
5 & 1145 & 13.83  \\
6 & 968 & 11.69  \\
7 & 1170 & 14.13  \\
8 & 944 & 11.40  \\
9 & 939 & 11.34  \\
10 & 916 & 11.06  \\
11 & 1055 & 12.74  \\
12 & 944 & 11.40  \\
\bottomrule
\end{tabular}}
\end{table}

\begin{table}[htbp]
\caption{Detail of FACES on DASS-21.}
\label{table_DASS_21}
\centering
\setlength{\tabcolsep}{1mm}{
\scriptsize
\begin{tabular}{c c c c c c}
\toprule
\textbf{Factor} & \textbf{Normal} & \textbf{Mild} & \textbf{Moderate} & \textbf{Severe} & \textbf{Extremely Severe} \\
\midrule
Depression & 6316 & 734 & 795 & 229 & 207 \\
Anxiety & 5358 & 643 & 1256 & 466 & 558 \\
Stress & 6807 & 668 & 494 & 243 & 69 \\
\bottomrule
\end{tabular}}
\end{table}

\subsection{Symptom Discrepancy Index}

In psychological assessments for emotion detection tasks, it is common practice to classify individuals into distinct states based on threshold values derived from total scores on standardized scales. For example, the DASS-21 scale comprises 21 items, with 7 specifically assessing depression symptoms. A total score exceeding 9 on these items typically indicates a depressive state, whereas scores of 9 or below suggest a healthy condition. Similar threshold-based classification methods are widely employed in scales such as the PHQ, HAMD, and SDS \cite{cuevas2000severity}.

However, relying solely on aggregated total scores for classification introduces significant challenges. Individuals with identical total scores can exhibit substantially different response patterns across individual items, reflecting distinct symptom profiles \cite{fried2015depression}. This variability complicates machine learning tasks by introducing inconsistencies within labeled groups, hindering the training of accurate predictive models.

Formally, let $\mathcal{D} = \{(x_j, A_j)\}_{j=1}^{N}$ denote a dataset where $x_j \in \mathbb{R}^n$ represents clinical or behavioral features of the $j$-th individual, and $A_j = (a_{1,j}, \dots, a_{m,j}) \in \mathbb{R}^m$ denotes their responses across $m$ scale items. The total score is computed as:
\begin{equation}
S_j = \sum_{i=1}^{m} a_{i,j},
\end{equation}
and the binary label is assigned as:
\begin{equation}
y_j = \begin{cases}
1, & \text{if } S_j > k,\\
0, & \text{otherwise}.
\end{cases}
\end{equation}
The probability of observing a label $y$ given feature $x$ is:

\begin{equation}
\begin{aligned}
P(y|x) = \int_{A \in \mathcal{A}_y} P(A|x)\, dA,
\\
\quad \text{with} \quad \mathcal{A}_y = \{A \in \mathbb{R}^m : g(A) = y\},
\end{aligned}
\end{equation}

illustrating how distinct item-level responses are aggregated under the same label, obscuring clinically significant heterogeneity.

The summation-based labeling approach simplifies clinical deployment but risks conflating patients with distinct symptom patterns sharing the same aggregate score. If two vectors $A_i$ and $A_{i'}$ differ but satisfy $\sum_{j=1}^m a_{ij} = \sum_{j=1}^m a_{i'j} > k$, both are labeled $y_i = y_{i'} = 1$. This surjective mapping masks meaningful variability and introduces label noise, as predictive models cannot distinguish between profiles with identical sums.

This label noise impacts model performance in two key ways:
\begin{enumerate}
    \item The label function $y_i = \mathbb{I}\bigl(\sum_{j=1}^m a_{ij} > k\bigr)$ generates misleading training signals, preventing classifiers from learning subtle clinical distinctions.
    \item Mislabeled or outlier data points distort the decision boundary, increasing model variance.
\end{enumerate}
Consequently, standard empirical risk minimization, $\frac{1}{N}\sum_{i=1}^{N} L(y_i, f(x_i; \theta))$, becomes less effective, elevating generalization error.

To address this limitation, we propose the \textit{Symptom Discrepancy Index}, a metric that quantifies variability in item-level responses among subjects with identical total scores, capturing inconsistencies arising from diverse symptom profiles.

SDI is computed in streamlined steps:
\begin{enumerate}
    \item \textbf{Construct and Standardize Item Response Vectors:}  
    For each subject, create a standardized item response vector:
    \begin{equation}
    z_{i,j} = \frac{a_{i,j} - \mu_i}{\sigma_i},
    \end{equation}
    Where \\
    $\quad \mu_i = \frac{1}{N}\sum_{j=1}^{N} a_{i,j}$, 
    $\quad \sigma_i = \sqrt{\frac{1}{N}\sum_{j=1}^{N} (a_{i,j} - \mu_i)^2}$.
    
    \item \textbf{Group Subjects by Total Score:}  
    Group subjects with the same total score $T$:
    \begin{equation}
    G_T = \{ j \mid S_j = T \}.
    \end{equation}
    
    \item \textbf{Compute Group Discrepancy Metrics:}  
    For each group $G_T$, calculate the mean vector $\bar{z}_i = \frac{1}{|G_T|}\sum_{j \in G_T} z_{i,j}$, then compute distances $d_j = \sqrt{\sum_{i=1}^{m} (z_{i,j} - \bar{z}_i)^2}$ for each $j \in G_T$. The group discrepancy index is:
    \begin{equation}
    \begin{aligned}
    GDI_{G_T} = \frac{1}{|G_T|}\sum_{j \in G_T} d_j +
    \\
    \sqrt{\frac{1}{|G_T|}\sum_{j \in G_T} \left( d_j - \frac{1}{|G_T|}\sum_{k \in G_T} d_k \right)^2}.
    \end{aligned}
    \end{equation}
    
    \item \textbf{Calculate SDI:}  
    Compute the overall SDI as the weighted average of $GDI_{G_T}$ across all groups:
    \begin{equation}
    SDI = \frac{\sum_{T} |G_T| \times GDI_{G_T}}{\sum_{T} |G_T|}.
    \end{equation}
\end{enumerate}
The SDI encapsulates both within-group variability and between-group differences, providing a comprehensive measure of symptom profile inconsistency.

To mitigate the effects of summation-based thresholding, we propose filtering observations with inconsistent item-level profiles. Let $P(A_j \mid y_j)$ represent the likelihood that $A_j$ corresponds to label $y_j$. Samples where $P(A_j \mid y_j) \leq \varepsilon$ are removed, yielding a filtered dataset $\mathcal{D}' = \mathcal{D} \setminus \{(x_j, A_j) \mid P(A_j \mid y_j) \leq \varepsilon\}$. Training on $\mathcal{D}'$ via:
\begin{equation}
\hat{\theta}' = \arg\min_{\theta \in \Theta} \frac{1}{|\mathcal{D}'|}\sum_{(x_i,y_i) \in \mathcal{D}'} L\bigl(y_i,f(x_i;\theta)\bigr),
\end{equation}
reduces label noise, decreasing estimation variance and improving the decision boundary. This results in a lower expected risk, $\mathbb{E}_{(x,y)\sim \mathcal{P}}[L(y,f(x;\hat{\theta}'))]$, compared to training on the unfiltered dataset.

In summary, while threshold-based summation is clinically convenient, it obscures item-level variability and introduces label noise. The SDI quantifies this heterogeneity, and filtering inconsistent samples enhances model performance by preserving clinically relevant distinctions.

\subsection{Emotional State Analysis}
In this section, to gain a deeper understanding of the interconnections among various emotional states such as depression, anxiety, and stress, we implemented advanced statistical analyses on the DASS-21 questionnaire data. We also introduced a novel factor that integrates these three emotional states and provided a data-driven categorization of emotional conditions.

\subsubsection{Correlation analysis}
The DASS-21 questionnaire utilized in our dataset consists of three self-report scales designed to assess the emotional states of depression, anxiety, and stress. It is important to note that these emotional states do not exist in isolation but often co-occur. To lay the groundwork for further analysis, we initiated our study with a correlational analysis to investigate the interrelations among depression, anxiety, and stress. Understanding the interplay among these factors is important to our research, as it can illuminate the complexities of emotional well-being. A detailed statistical analysis of these factors has been conducted to extract deeper insights into their dynamics.

To explore the predictive dynamics among these emotional states, we initially examined the relationships between depression, anxiety, and stress using Pearson's correlation, aiming to determine the extent of linear association among these variables. Next, we used linear regression analysis to demonstrate how variations in depression, anxiety, and stress levels can predict each other, unveiling directional interactions among these emotional states. This approach reveals their intricate interdependencies and illuminates the mechanisms underpinning their relationships, providing deeper insights into their interconnectedness. The regression equation was simplified as follows:

\begin{align}
Y_i = \alpha_0 + \sum_{j=1}^{3}\alpha_j X_{ij} + \sum_{j=1}^{3}\sum_{k=1, k \neq j}^{3}\beta_{jk} X_{ij}X_{ik} + \epsilon_i
\end{align}
Here, $Y_i$ represents the dependent variable corresponding to one of the factors. $X_{ij}$ denotes the influence of the $j$-th factor on the $i$-th factor. $\alpha_0$ is the intercept of the model, indicating the baseline level when all factors are zero. $\alpha_j$ are the coefficients that reflect the direct impact of the $j$-th factor on the $i$-th factor. $\beta_{jk}$ are the interaction coefficients, capturing the combined effect of the $j$-th and $k$-th factors on the $i$-th factor. $\epsilon_i$ is the error term, accounting for the variability not explained by the model.

\subsubsection{Clustering analysis}

The traditional approach to analyzing the DASS-21 scale, by categorizing depression, anxiety, and stress as separate entities, fails to acknowledge their intrinsic interrelation, potentially leading to distorted interpretations. Moreover, the application of arbitrary thresholds to separate emotional states inadequately reflects the spectrum of individual variations, thus diminishing the scale's applicability to a broad array of demographic groups. Therefore, we employed clustering methods as an innovative solution, facilitating a data-driven examination of the DASS-21 dataset. This approach uncovers organic clusters that reflect the complex relationships among emotional states, overcoming the constraints of traditional analysis and generic thresholds. 

We employed Agglomerative Clustering to identify subgroups within the DASS-21 scales dataset. To ensure the effectiveness of clustering, we use Silhouette Coefficient, Davies-Bouldin Index and Calinski-Harabasz Index to evaluate the cluster performance.

\subsection{Facial Feature Extraction}
We employed OpenFace2.0 \cite{baltrusaitis_openface_2018}, a facial behavior analysis toolkit, for a comprehensive extraction of facial features from video sequences. We extracted a total of 8,508 features from each video, setting the foundation for subsequent analysis and modeling. The methodology is structured as follows:

\subsubsection{Video Preprocessing}
\begin{itemize}
\item Face Detection: OpenFace2.0 employs a face detection algorithm based on Multi-task Convolutional Neural Networks (MTCNN) \cite{zhang2016joint}. This algorithm, trained on many large face datasets, enhances the accuracy of face detection, particularly excelling in handling side faces and highly occluded faces.
\item Facial Landmark Detection and Tracking: The tool then utilizes the Convolutional Experts Constrained Local Model (CE-CLM) \cite{zadeh2017convolutional} for detecting and tracking facial landmarks. CE-CLM combines a Point Distribution Model (PDM) with local appearance models to accurately capture the variations in facial landmarks. Optimizations for real-time performance, incorporating techniques like model simplification, intelligent multi-hypothesis selection, and sparse response map calculation.
\end{itemize}

\subsubsection{Feature Extraction}
Firstly, we extracted image features by downsampling the preprocessed videos from 30 fps to 1 fps. Subsequently, we extracted frame-by-frame sequences of face-aligned images, which served as the \textbf{Sequential Image} data for the subsequent deep learning models.

\begin{figure}[htbp]
\centering
\includegraphics[width=\columnwidth]{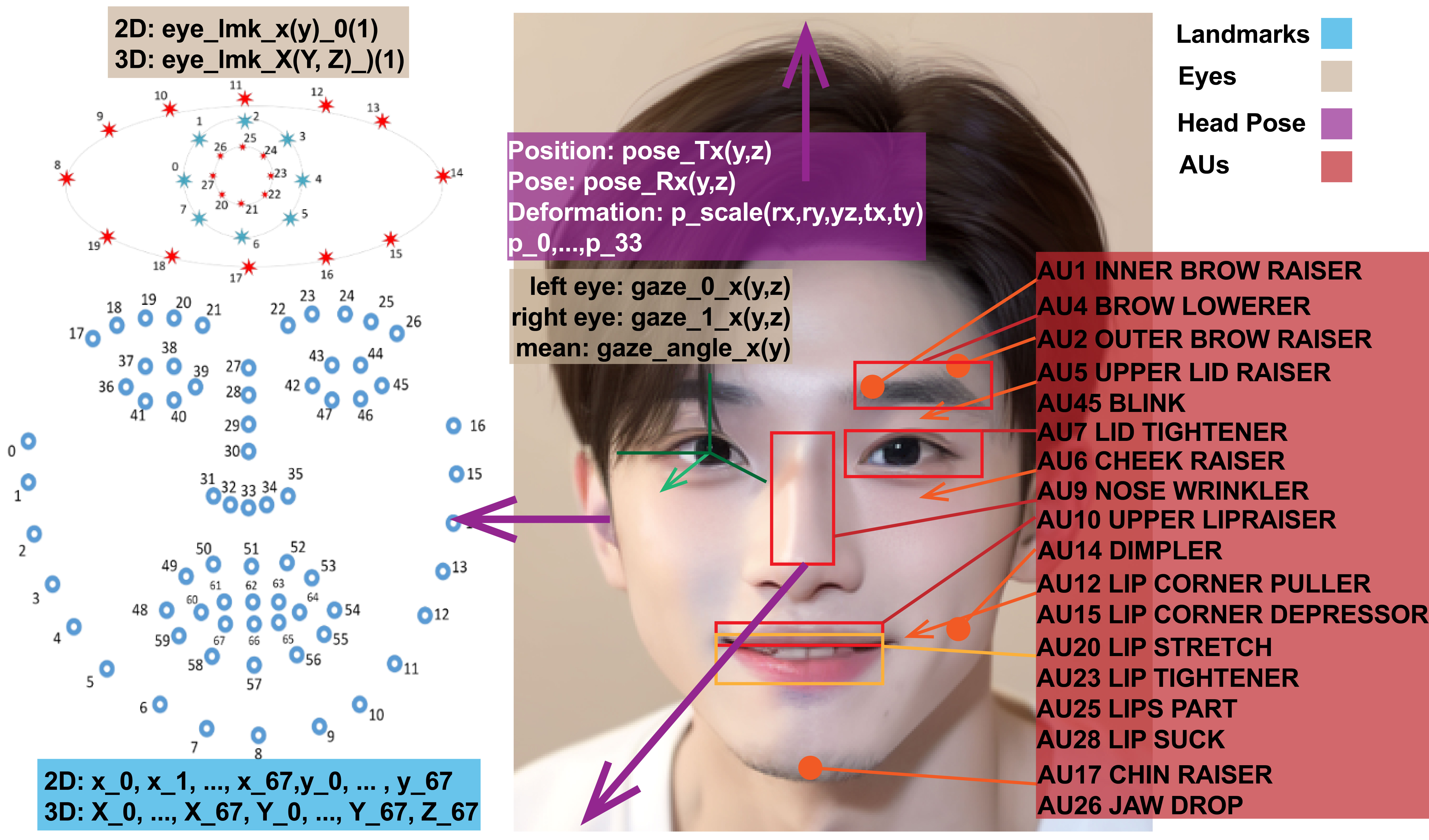}
\caption{\textbf{Introduction of feature labels related facial images.}}
\label{fig2}
\end{figure}

Next, we extract \textbf{Statistical Features} with OpenFace2.0. Each of these primary features provides us with a wealth of information about an individual’s emotions, intentions, and reactions. By holistically interpreting them, we can gain a better understanding of underlying sentiments and behavioral nuances. Detailed association with facial images of them is illustrated in \textbf{Figure} \textbf{\ref{fig2}}.

\subsubsection{Post-processing and Feature Aggregation}
To achieve a consistent feature representation across videos with varied frame counts, we aggregated statistical features from sequential frames into lists. For every list, we computed:

\textbf{(1) Central Tendency:} Mean and Median; \textbf{(2) Dispersion:} Maximum, Minimum, Standard Deviation, Variance, Range, 25th percentile, 75th percentile, and Interquartile Range; \textbf{(3) Shape:} Skewness and Kurtosis; \textbf{(4) Derivative:} Delta and Delta-Delta (Delta is differentiation).

These metrics offered insights into the dynamic variations of facial features. We finally got 8508 statistical features.

\subsection{Machine Learning Models}

We selected a diverse set of models specifically designed for mental disorder detection, with a focus on their ability to effectively analyze both statistical features and facial expression video. Specifically, machine learning models process numerical statistical features and deep learning models process sequential image data. The aim was to evaluate models capable of capturing the complex relationships between facial cues and mental health conditions, ensuring robust performance across both classification and regression tasks. The detailed definitions of these models and the experimental design are as follows:

\subsubsection{Model Definition}

\begin{itemize}
\item \textbf{SVM:} Support Vector Machine implemented using the Scikit-learn library \cite{pedregosa2011scikit}. We employed an RBF kernel with hyperparameters optimized via grid search. The parameter grid included the regularization parameter $C \in \{0.1, 1, 10, 100\}$ and the kernel coefficient $\gamma \in \{10^{-3}, 10^{-2}, 10^{-1}, 1\}$. Five-fold cross-validation was utilized to evaluate the performance of each parameter combination, selecting the model that minimized the cross-validation error within the specified threshold.

\item \textbf{Random Forest:} Random Forest also implemented with Scikit-learn. The model was configured with 1000 decision trees to ensure robustness and mitigate overfitting. Hyperparameter tuning was conducted using grid search over the following parameters: number of features considered for splitting $max\_features \in \{\text{auto}, \sqrt{n\_features}, \log_2(n\_features)\}$, maximum tree depth $max\_depth \in \{10, 50, 100, \text{None}\}$, and the minimum number of samples required to split an internal node $min\_samples\_split \in \{2, 5, 10\}$. A five-fold cross-validation approach was applied to identify the optimal combination of hyperparameters that minimized prediction errors.

\item \textbf{LI-FPN:} LI-FPN is designed for the automatic detection of depression and anxiety disorders from facial video frames \cite{li2023li}. It consists of two primary components: the Learning and Imitation Module (LIM) and the Spatio-temporal Feature Pyramid Network (STFPN). LIM uses an attention mechanism to enhance feature extraction through ``learning" and ``imitation" phases. The attention map AM$i$ is created as follows:

\begin{equation}
\text{AM}i = \text{Sigmoid}\left(\sum{t=1}^T \text{Sigmoid}(\text{Conv}(\phi{t}^{\text{BB}i}))\right)
\end{equation}

STFPN integrates spatial and temporal features using a pyramid structure. The fused features $\phi^{\text{DC}}$ are obtained by:

\begin{equation}
\phi^{\text{DC}} = \text{DownSample}(\text{Conv}k(\phi{m-1}^{\text{f}})) + \phi_{m+1}^{\text{LI}}
\end{equation}

LI-FPN demonstrates high accuracy in detecting depression and anxiety, leveraging complex interactions between spatial and temporal data.

\item \textbf{MSN:} Multiscale Spatiotemporal Network is a deep learning model designed for automatic depression recognition from facial videos \cite{de2020deep}. It employs a 3D Convolutional Neural Network (CNN) architecture to capture both spatial and temporal information. The core of MSN is composed of multiple parallel convolutional layers with varying temporal depths and receptive fields, enabling the model to effectively capture a wide range of facial dynamics associated with depressive behaviors. The output of the basic building block is given by:

\begin{equation}
\hat{y} = \sigma(\text{BN}(H(x, \{H_i\}_{i=1}^M)) + x)
\end{equation}

Where $H(x)$ represents the residual mapping function, and $H_i$ are the parameters of the convolutional layers. Experimental results show that MSN outperforms state-of-the-art methods in automatic depression recognition.

\item \textbf{MTDAN:} MTDAN is designed for the multi-task detection of anxiety and depression through voice analysis \cite{zhang2023mtdan}. It consists of two main components: the Multi-Task Learning Network (MTLN) and the Adaptive Fusion Module (AFM). MTLN employs a series of task-specific sub-networks to simultaneously predict anxiety and depression levels, with each network being optimized using a shared backbone. The output features $\phi^{\text{MTLN}}$ are obtained by:

\begin{equation}
\phi^{\text{MTLN}} = \sum_{i=1}^{N} \text{Sigmoid}\left(\text{Conv}(\phi_i^{\text{Input}})\right)
\end{equation}

AFM is responsible for fusing the task-specific features using a weighted attention mechanism to optimize task alignment. The fused features $\phi^{\text{AFM}}$ are obtained as follows:

\begin{equation}
\phi^{\text{AFM}} = \sum_{i=1}^{N} \alpha_i \cdot \phi_i^{\text{MTLN}}
\end{equation}

Where $\alpha_i$ is the weight determined by the attention mechanism. MTDAN achieves state-of-the-art performance in both anxiety and depression detection by effectively leveraging the interdependencies between the two tasks.

\end{itemize}

\subsubsection{Experimental Setup}

We assess the performance of our dataset using two types of tasks: regression and classification, employing both statistical features and sequential facial images. For both tasks, we utilize traditional machine learning models for statistical features, and deep learning models for sequential frame image features. In classification tasks involve predicting mental health disorders using DASS-21 scale divisions and a new factor Cluster.

All tasks follow a systematic cross-validation approach: five-fold cross-validation to ensure the robustness and generalisability of machine learning models. For deep learning tasks, the dataset is split into training, validation, and test segments following a 7:2:1 proportion.

For the regression tasks, the primary evaluation metrics employed are Mean Absolute Error (MAE) and Root Mean Square Error (RMSE):

For the classification tasks, we utilize several standard evaluation metrics, including Accuracy, F1 score, Precision, Recall, and AUC. A grid search is employed for hyperparameter tuning, including the learning rate, batch size, and number of epochs.

\subsubsection{Loss Functions}

For deep learning models, we use L1 Loss for regression tasks and Cross-Entropy Loss for classification tasks.

\subsubsection{Implementation Details}

All machine learning models were trained using the scikit-learn library \cite{pedregosa2011scikit} and all deep learning models were trained using the Adam optimizer on a Tesla A100 GPU with 40GB of memory.

For both regression and classification tasks, we employed an extensive hyperparameter tuning process to optimize the performance of each model. This process involved adjusting various parameters, including but not limited to learning rate, batch size, number of epochs, and model architecture.

We used the Adam optimizer for all deep learning models, with learning rates typically ranging from $1e^{-5}$ to $1e^{-2}$. Batch sizes were varied between 8 and 32, depending on the specific task and available memory. The number of training epochs was determined dynamically for each model, with early stopping mechanisms implemented to prevent overfitting.

The final hyperparameters for each model were selected based on their performance on a validation set, ensuring the best possible generalization to unseen data. This meticulous tuning process allowed us to adapt our models to the specific characteristics of each task, resulting in optimized performance across all experiments.

\section{Experimental Results}

The results section is organized around four successive analyses. First, we characterize the FACES corpus by mapping the distribution of total DASS‑21 scores and benchmarking it against public datasets. Second, we quantified item‑level heterogeneity with the Group Discrepancy Index and Symptom Discrepancy Index, complemented by correlation and clustering analyses of depression, anxiety, and stress. Third, we implemented binary and refined binary classification pipelines - using random forests, support vector machines, LI-FPN, MSN, and MTDAN - to differentiate normal from abnormal psychological states under varying filtering thresholds based on GDI. Finally, we performed continuous regression of each symptom domain, again contrasting full‑corpus training with models trained on filtered subsets.

\subsection{Symptom Heterogeneity in the FACES Dataset}

To examine the symptom variability embedded in the datasets, we adopted several complementary perspectives: first, the overall distribution of total scores; next, the item-level variability within iso-score groups; and finally, a combined analysis of inter-symptom relations and data-driven clustering. Each analysis was conducted separately for the depression, anxiety, and stress sub-scales of the DASS-21, thereby enabling direct comparison across these symptom dimensions.

\subsubsection{Total Score Distribution across corpora}

\begin{figure}[htbp]
\centering
\includegraphics[width=\columnwidth]{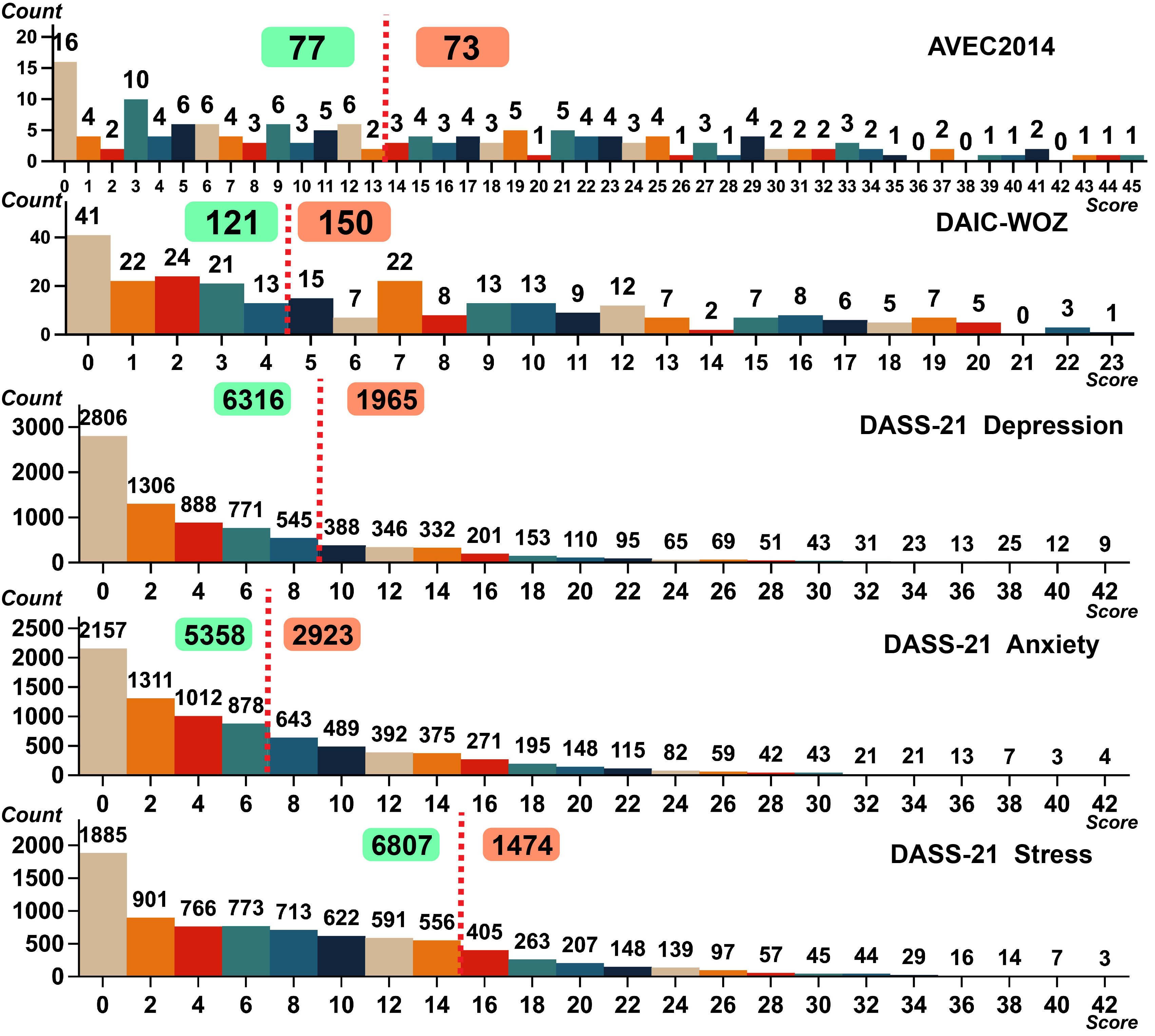}
\caption{\textbf{Distribution of total Scores in AVEC2014, DAIC-WOZ, and FACES Datasets. The orange dotted line represents the clinical cutoff for mental health disorders, dividing participants into “normal” (left of the line) and “abnormal” (right of the line) psychological states.}}
\label{fig3}
\end{figure}

To analyze the heterogeneity of the datasets, we first examined the distribution of total scores from the scales. \textbf{Figure \ref{fig3}} compares the histograms of total scores gathered with the DASS-21 (FACES), BDI-II (AVEC-2014) and PHQ-8 (DAIC-WOZ). In line with large-scale population-based studies \cite{steel2014global}, the FACES distribution mainly contains participants whose scores fall below the clinical cut-offs for depression, anxiety and stress (n = 6316, 5358 and 6807, respectively). By contrast, AVEC-2014 and DAIC-WOZ show roughly symmetric distributions around the cut-off scores, a feature that makes model training easier but puts too much weight on severe symptoms compared with their frequency in the general population. Therefore, the right-skewed shape seen in FACES offers a more realistic evaluation set for algorithms that will be used in practical screening settings, where uneven class sizes are common.

\begin{figure*}[htbp]
\centering
\includegraphics[width=2\columnwidth]{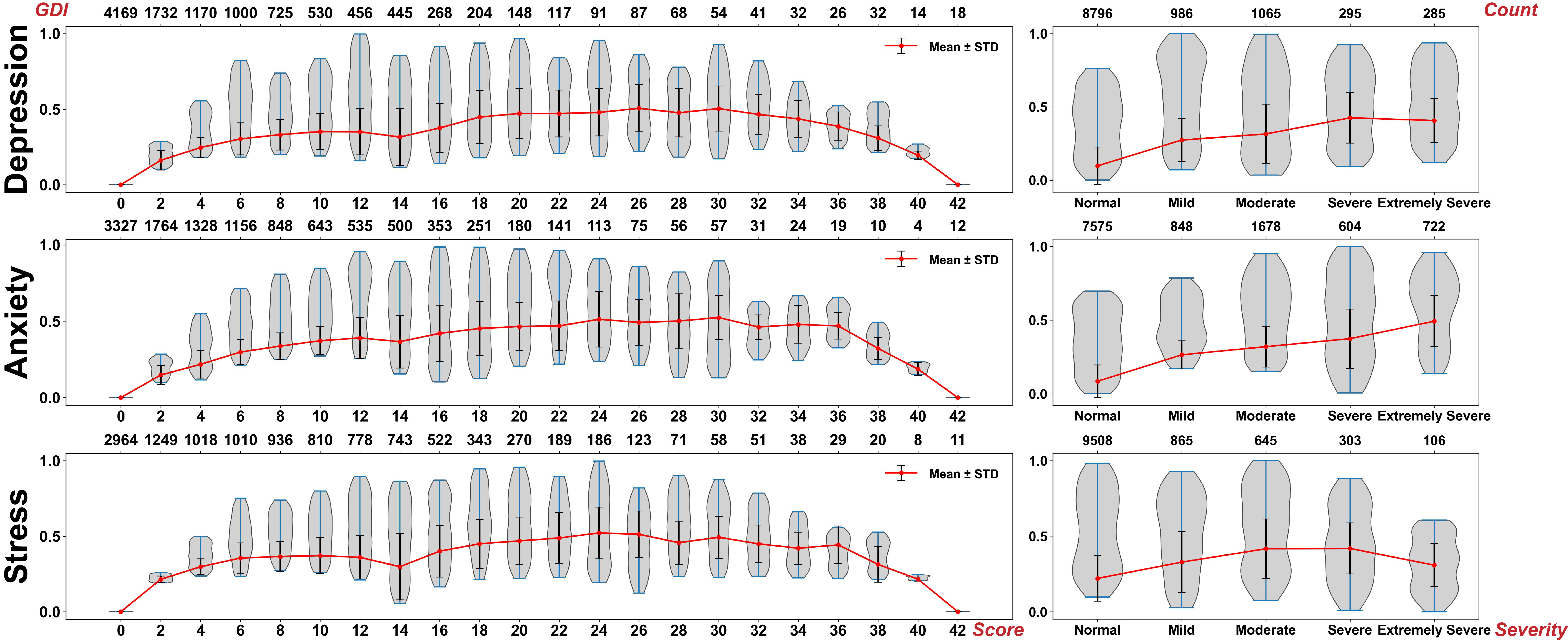}
\caption{\textbf{Distribution of Groups with Same Total Score and Same Severity.}}
\label{fig4}
\end{figure*}

\subsubsection{Item‑level heterogeneity at fixed total scores}

Although the total score is frequently treated as a adequate summary measure, individuals who achieve the same total score may select very different combinations of items. This phenomenon is visualized for FACES in \textbf{Figure \ref{fig4}} (left panels), where violin plots depict the distribution of Group Discrepancy Index ($GDI$) values across the entire range of possible totals (0–42). Heterogeneity is minimal at the extremes scores where either no symptom is present or nearly all items are selected, and reaches its highest level in the middle range (scores 10–22). The inverted U-shaped pattern is consistent across the depression, anxiety and stress sections, indicating that the variety of symptom combinations peaks at moderate severity. From a machine‑learning perspective, this non‑monotonic relationship implies that observations located in the diagnostically most ambiguous region also exhibit the greatest internal variability, complicating feature learning and decision‑boundary placement.

\begin{figure}[htbp]
\centering
\includegraphics[width=0.8\columnwidth]{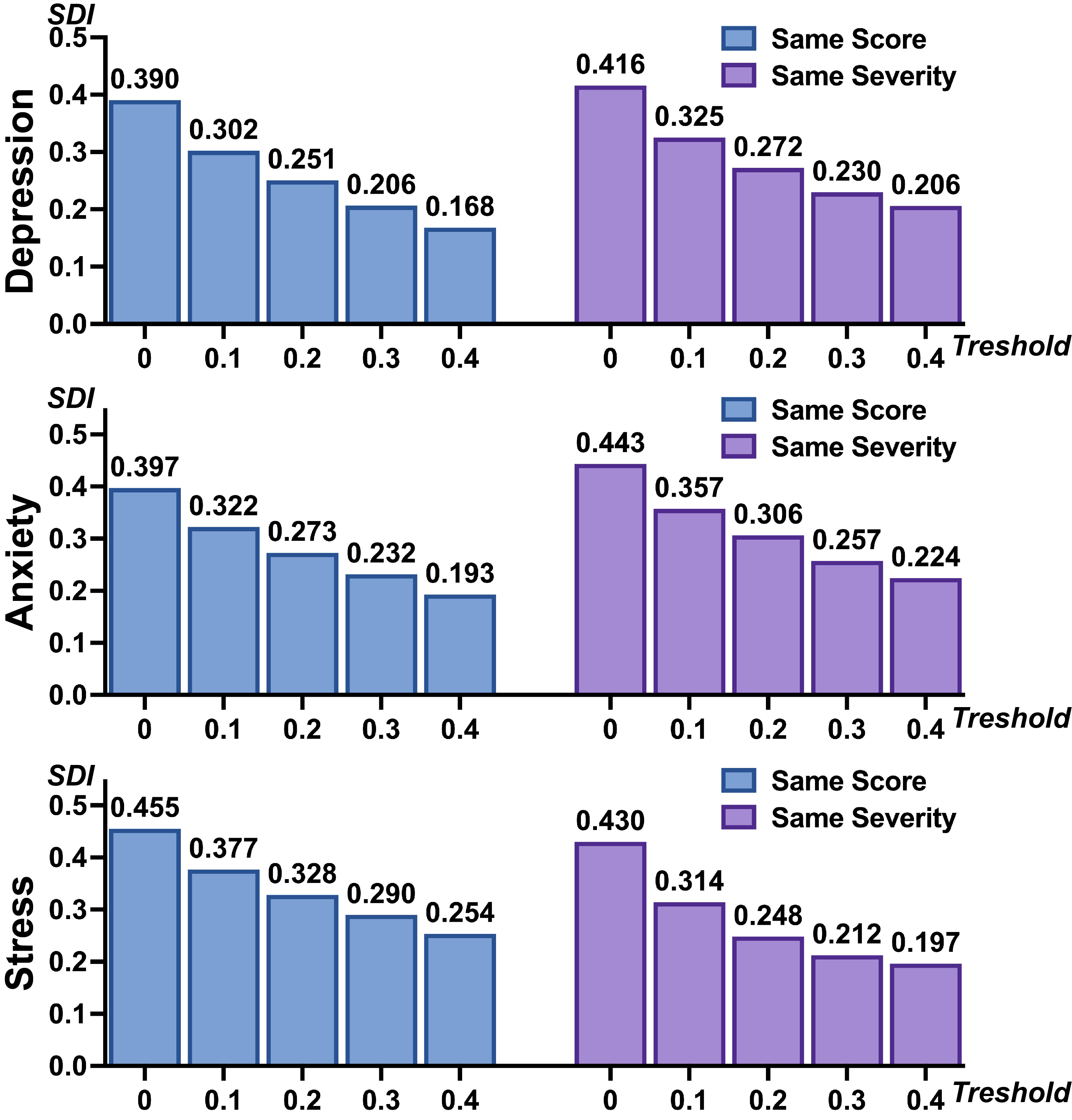}
\caption{\textbf{Same Score and Severity's SDI of Different Exclusion Thresholds.}}
\label{fig5}
\end{figure}

Furthermore, to examine heterogeneity that persists after coarse categorization, participants were reassigned to the five conventional DASS severity bands. For each band, SDI was calculated before and after iterative removal of the most discrepancies 5\% to 40\% of cases (\textbf{Figure \ref{fig5}}). Baseline SDI values confirm substantial residual heterogeneity even when the total score is translated into a categorical diagnosis (depression = 0.390, anxiety = 0.370, stress = 0.455). As the exclusion threshold increases, the SDI values gradually decrease, indicating reduced heterogeneity among the samples under stricter filtering conditions. These findings are consistent with dimensional models of psychopathology \cite{trull2005categorical}, which posit that categorical thresholds obscure continuous and highly individualized symptom topographies.

\subsubsection{Inter‑symptom correlations and latent cluster structure}

\begin{figure}[htbp]
\centering
\includegraphics[width=\columnwidth]{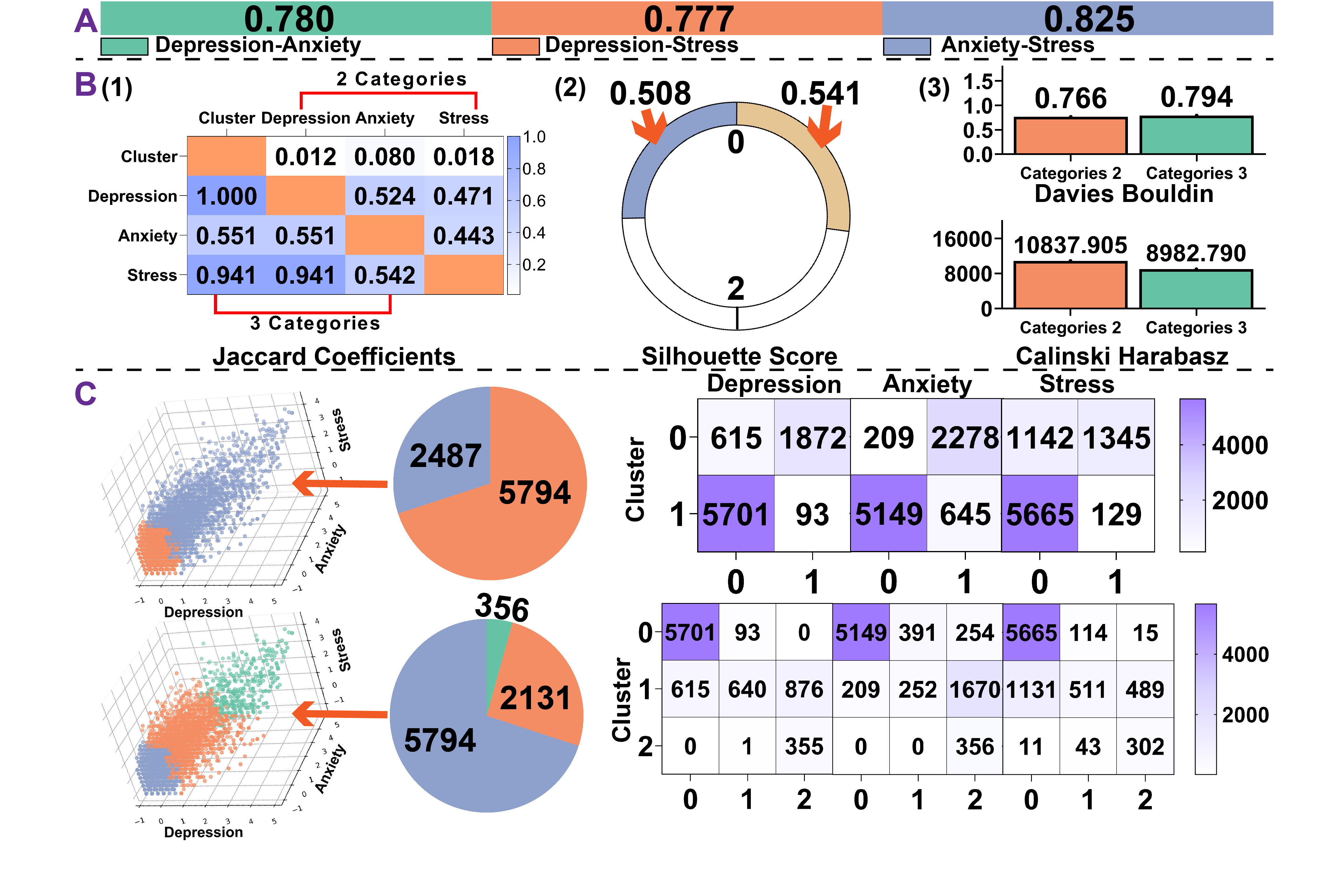}
\caption{\textbf{Overview of Correlation Analysis and Clustering Results.} \textbf{A} shows the correlation coefficients between depression, anxiety, and stress. \textbf{B} presents the Jaccard coefficients for two- and three-category cluster distinctions, along with silhouette scores, Davies-Bouldin Index, and Calinski-Harabasz Index for different cluster categorizations. \textbf{C} illustrates the clustering results, indicating the distribution of individuals into distinct groups based on symptom severity.}
\label{fig6}
\end{figure}

To better understand the dataset, we analyzed the correlations among depression, anxiety, and stress. \textbf{Table \ref{table_factor_correlation}} and \textbf{Figure \ref{fig6}A} document robust positive correlations among the three symptom domains, with anxiety–stress exhibiting the largest coefficient (r=0.825), followed by depression–anxiety (r=0.780) and depression–stress (r=0.777). Linear regressions confirmed that the variance in one symptom class could be explained by another, reinforcing the conceptualization of these constructs as partially overlapping manifestations of a broader internalizing spectrum.

To investigate the latent structure emerging from these interdependencies, K‑means clustering was performed in the three‑dimensional symptom space. The comparative validity metrics (\textbf{Table \ref{table_cluster_analysis}} and  \textbf{Figure \ref{fig6}B}) favored a two‑cluster solution over a three‑cluster alternative: the silhouette coefficient increased from 0.508 to 0.541, the Davies–Bouldin index decreased from 0.794 to 0.766, and the Calinski–Harabasz index rose from 8982.790 to 10837.905. Cluster0 (n=2487) aggregated individuals exhibiting clinically significant elevations on at least one sub‑scale, whereas Cluster1 (n=5794) corresponded to asymptomatic or minimally symptomatic participants (\textbf{Figure \ref{fig6}C}). The relative sizes of these clusters paralleled the prevalence estimates derived from the univariate cut‑offs, lending external validity to the unsupervised partitioning.

\begin{table}[htbp]
\renewcommand{\arraystretch}{1.1}
\caption{Correlation and Regression Analysis of Inter-Factors.}
\label{table_factor_correlation}
\centering
\setlength{\tabcolsep}{2pt}
\footnotesize
\begin{threeparttable}
\begin{tabular}{c c c c c}
\toprule
\textbf{Factor} & \textbf{Correlated With} & \textbf{$r$} & \textbf{$R^2$} & \textbf{$\beta$}\\
\midrule
Depression & Anxiety & \multirow{2}*{0.780} & \multirow{2}*{60.9\%} & 0.816  \\
Anxiety & Depression & & & 0.746 \\
\midrule
Anxiety & Stress & \multirow{2}*{0.825} & \multirow{2}*{68.0\%} & 0.750  \\
Stress & Anxiety & & & 0.907  \\
\midrule
Depression & Stress & \multirow{2}*{0.777} & \multirow{2}*{60.4\%} & 0.739 \\
Stress & Depression & & & 0.817 \\
\bottomrule
\end{tabular}
\begin{tablenotes}
\item* $r$, $R^2$, $\beta$ represent Correlation, Variance Explained and Regression Coefficient, respectively.
\end{tablenotes}
\end{threeparttable}
\end{table}

\begin{table}[htbp]
\renewcommand{\arraystretch}{1.1}
\caption{Cluster Analysis Summary.}
\label{table_cluster_analysis}
\centering
\setlength{\tabcolsep}{8pt}
\footnotesize
\begin{threeparttable}
\begin{tabular}{c c c c}
\toprule
\textbf{N-Clusters} & \textbf{SC} & \textbf{DB} & \textbf{CH} \\
\midrule
$N$ = 2 & 0.541 & 0.766 & 10,837.905 \\
$N$ = 3 & 0.508 & 0.794 & 8,982.790 \\
\bottomrule
\end{tabular}
\begin{tablenotes}
\item* SC, DB, CH represent Silhouette Coefficient, Davies-Bouldin Index and Calinski-Harabasz Index, respectively.
\end{tablenotes}
\end{threeparttable}
\end{table}

\subsection{Classification Performance}

The predictive experiments used the three DASS‑21 sub‑scores, Depression (D), Anxiety (A) and Stress (S), together with the data‑driven Cluster label (C). Two binary scenarios were considered. In the standard binary classification (BC) setting each record was tagged as Normal or Abnormal; the latter combined Mild, Moderate, Severe and Extremely‑Severe cases, while the cluster label was taken from the two‑cluster solution described earlier. In the refined binary classification (RBC) setting the Mild level was removed in order to sharpen the boundary between health and disorder; the remaining classes were Normal and Moderate‑to‑Extremely‑Severe. For the cluster variable the three‑cluster partition was used, with the middle group omitted to preserve a binary structure.

When the full FACES corpus was used without any filtering, every classifier returned accuracies between 0.55 and 0.60 (\textbf{Table \ref{table_bc_gdi}} and \textbf{Table \ref{table_rbc_gdi}}). These moderate values confirm that wide symptom overlap and subjective cut‑offs greatly hinder model learning.

To reduce this effect, we excluded the 10\% of samples that lay farthest from their group centres, as identified by the Group Discrepancy Index (GDI). This filtering step improved performance for every model. In the BC task, the SVM’s F1 score for the Cluster label rose from 0.523 to 0.640, while the MTDAN’s F1 climbed from 0.602 to 0.637. The gains were even larger in the RBC task: the RF’s F1 for Cluster increased from 0.575 to 0.786, and the SVM’s F1 reached 0.779. Both MSN and MTDAN showed similar improvements; MTDAN, for example, achieved 0.812 accuracy for Cluster. The stronger boost in RBC confirms that the Mild category adds substantial noise, which can hide meaningful patterns.

With respect to model families, RF provided a solid baseline thanks to its robustness against over‑fitting and its ability to handle non‑linear decision surfaces. SVM excelled when the filtered data offered a clearer margin, as seen in its RBC results. The two deep networks performed competitively once the training set was purged of ambiguous cases, and MTDAN benefited further from its multi‑task design, which exploits shared structure across labels.

\begin{table*}[!ht]
\renewcommand{\arraystretch}{1}
\caption{BC Performance Metrics Across Different GDI Thresholds}
\label{table_bc_gdi}
\centering
\setlength{\tabcolsep}{1.8pt}
\footnotesize
\begin{tabular}{c c | c c c c | c c c c}
\toprule
\multirow{2}{*}{\textbf{Method}} & \multirow{2}{*}{\textbf{Metric}} & \multicolumn{4}{c|}{\textbf{Threshold = 0}} & \multicolumn{4}{c}{\textbf{Threshold = 0.1}} \\
\cmidrule(lr){3-6} \cmidrule(lr){7-10}
~ & ~ & \textbf{D} & \textbf{A} & \textbf{S} & \textbf{C} & \textbf{D} & \textbf{A} & \textbf{S} & \textbf{C} \\
\midrule
\multirow{4}{*}{\textbf{RF}} 
& ACC       & 0.550±0.09 & 0.568±0.07 & 0.570±0.06 & 0.583±0.05 & 0.617±0.04 & 0.626±0.03 & 0.639±0.02 & 0.642±0.01 \\
~ & Precision & 0.557±0.08 & 0.561±0.06 & 0.589±0.05 & 0.590±0.04 & 0.623±0.03 & 0.620±0.02 & 0.649±0.01 & 0.659±0.00 \\
~ & Recall    & 0.528±0.12 & 0.530±0.09 & 0.523±0.08 & 0.531±0.07 & 0.592±0.06 & 0.603±0.05 & 0.611±0.04 & 0.622±0.03 \\
~ & F1        & 0.532±0.10 & 0.546±0.08 & 0.550±0.07 & 0.563±0.06 & 0.601±0.05 & 0.612±0.04 & 0.620±0.03 & 0.632±0.02 \\
\midrule
\multirow{4}{*}{\textbf{SVM}} 
& ACC       & 0.551±0.09 & 0.513±0.15 & 0.545±0.07 & 0.552±0.06 & 0.594±0.05 & 0.604±0.04 & 0.612±0.03 & 0.625±0.02 \\
~ & Precision & 0.559±0.08 & 0.513±0.18 & 0.559±0.06 & 0.566±0.05 & 0.628±0.04 & 0.634±0.03 & 0.649±0.02 & 0.655±0.01 \\
~ & Recall    & 0.517±0.14 & 0.442±0.25 & 0.485±0.09 & 0.496±0.08 & 0.583±0.07 & 0.621±0.06 & 0.630±0.05 & 0.644±0.04 \\
~ & F1        & 0.529±0.11 & 0.471±0.20 & 0.514±0.08 & 0.523±0.07 & 0.603±0.06 & 0.630±0.05 & 0.634±0.04 & 0.640±0.03 \\
\midrule
\multirow{4}{*}{\textbf{LI-FPN}} 
& ACC       & 0.573±0.07 & 0.581±0.06 & 0.598±0.05 & 0.602±0.04 & 0.592±0.03 & 0.598±0.02 & 0.616±0.01 & 0.622±0.00 \\
~ & Precision & 0.576±0.08 & 0.587±0.07 & 0.603±0.06 & 0.613±0.05 & 0.608±0.04 & 0.606±0.03 & 0.625±0.02 & 0.631±0.01 \\
~ & Recall    & 0.540±0.13 & 0.554±0.09 & 0.544±0.08 & 0.555±0.07 & 0.575±0.06 & 0.587±0.05 & 0.594±0.04 & 0.604±0.03 \\
~ & F1        & 0.538±0.10 & 0.569±0.08 & 0.575±0.07 & 0.583±0.06 & 0.583±0.05 & 0.591±0.04 & 0.601±0.03 & 0.611±0.02 \\
\midrule
\multirow{4}{*}{\textbf{MSN}} 
& ACC       & 0.589±0.06 & 0.600±0.05 & 0.614±0.04 & 0.625±0.03 & 0.610±0.02 & 0.622±0.01 & 0.630±0.00 & 0.639±0.02 \\
~ & Precision & 0.590±0.07 & 0.605±0.06 & 0.625±0.05 & 0.638±0.04 & 0.626±0.03 & 0.621±0.02 & 0.642±0.01 & 0.651±0.00 \\
~ & Recall    & 0.565±0.11 & 0.573±0.09 & 0.568±0.08 & 0.571±0.07 & 0.593±0.06 & 0.604±0.05 & 0.610±0.04 & 0.622±0.03 \\
~ & F1        & 0.578±0.09 & 0.589±0.08 & 0.594±0.07 & 0.606±0.06 & 0.606±0.05 & 0.617±0.04 & 0.624±0.03 & 0.632±0.02 \\
\midrule
\multirow{4}{*}{\textbf{MTDAN}} 
& ACC       & 0.585±0.07 & 0.597±0.06 & 0.610±0.05 & 0.621±0.04 & 0.615±0.03 & 0.627±0.02 & 0.635±0.01 & 0.644±0.00 \\
~ & Precision & 0.587±0.08 & 0.602±0.07 & 0.621±0.06 & 0.634±0.05 & 0.631±0.04 & 0.626±0.03 & 0.647±0.02 & 0.656±0.01 \\
~ & Recall    & 0.562±0.12 & 0.570±0.09 & 0.565±0.08 & 0.568±0.07 & 0.598±0.06 & 0.609±0.05 & 0.615±0.04 & 0.627±0.03 \\
~ & F1        & 0.574±0.10 & 0.586±0.08 & 0.591±0.07 & 0.602±0.06 & 0.611±0.05 & 0.622±0.04 & 0.629±0.03 & 0.637±0.02 \\
\bottomrule
\end{tabular}
\end{table*}

\begin{table*}[!ht]
\renewcommand{\arraystretch}{1}
\caption{RBC Performance Metrics Across Different GDI Thresholds}
\label{table_rbc_gdi}
\centering
\setlength{\tabcolsep}{1.8pt}
\footnotesize
\begin{tabular}{c c | c c c c | c c c c}
\toprule
\multirow{2}{*}{\textbf{Method}} & \multirow{2}{*}{\textbf{Metric}} & \multicolumn{4}{c|}{\textbf{Threshold = 0}} & \multicolumn{4}{c}{\textbf{Threshold = 0.1}} \\
\cmidrule(lr){3-6} \cmidrule(lr){7-10}
~ & ~ & \textbf{D} & \textbf{A} & \textbf{S} & \textbf{C} & \textbf{D} & \textbf{A} & \textbf{S} & \textbf{C} \\
\midrule
\multirow{4}{*}{\textbf{RF}} 
& ACC       & 0.568±0.08 & 0.570±0.07 & 0.583±0.06 & 0.591±0.05 & 0.754±0.04 & 0.725±0.03 & 0.690±0.02 & 0.766±0.01 \\
~ & Precision & 0.560±0.09 & 0.575±0.06 & 0.597±0.05 & 0.605±0.04 & 0.756±0.03 & 0.726±0.02 & 0.749±0.01 & 0.768±0.00 \\
~ & Recall    & 0.534±0.12 & 0.547±0.08 & 0.538±0.07 & 0.547±0.06 & 0.792±0.05 & 0.785±0.04 & 0.698±0.03 & 0.807±0.02 \\
~ & F1        & 0.547±0.10 & 0.552±0.09 & 0.565±0.08 & 0.575±0.07 & 0.779±0.06 & 0.755±0.05 & 0.728±0.04 & 0.786±0.03 \\
\midrule
\multirow{4}{*}{\textbf{SVM}} 
& ACC       & 0.545±0.09 & 0.524±0.15 & 0.550±0.07 & 0.560±0.06 & 0.705±0.05 & 0.713±0.04 & 0.694±0.03 & 0.778±0.02 \\
~ & Precision & 0.554±0.08 & 0.517±0.18 & 0.569±0.06 & 0.573±0.05 & 0.756±0.04 & 0.829±0.03 & 0.836±0.02 & 0.849±0.01 \\
~ & Recall    & 0.523±0.13 & 0.453±0.20 & 0.491±0.09 & 0.504±0.08 & 0.698±0.07 & 0.879±0.06 & 0.821±0.05 & 0.901±0.04 \\
~ & F1        & 0.547±0.11 & 0.552±0.16 & 0.565±0.07 & 0.575±0.06 & 0.779±0.05 & 0.755±0.04 & 0.728±0.03 & 0.786±0.02 \\
\midrule
\multirow{4}{*}{\textbf{LI-FPN}} 
& ACC       & 0.586±0.07 & 0.595±0.06 & 0.604±0.05 & 0.616±0.04 & 0.754±0.03 & 0.705±0.02 & 0.668±0.01 & 0.733±0.00 \\
~ & Precision & 0.589±0.08 & 0.605±0.07 & 0.597±0.06 & 0.621±0.05 & 0.756±0.04 & 0.829±0.03 & 0.714±0.02 & 0.734±0.01 \\
~ & Recall    & 0.555±0.10 & 0.565±0.09 & 0.554±0.08 & 0.564±0.07 & 0.781±0.06 & 0.583±0.12 & 0.671±0.05 & 0.769±0.04 \\
~ & F1        & 0.589±0.09 & 0.625±0.08 & 0.585±0.07 & 0.590±0.06 & 0.765±0.05 & 0.739±0.04 & 0.695±0.03 & 0.752±0.02 \\
\midrule
\multirow{4}{*}{\textbf{MSN}} 
& ACC       & 0.609±0.06 & 0.610±0.05 & 0.627±0.04 & 0.635±0.03 & 0.746±0.02 & 0.718±0.01 & 0.687±0.00 & 0.752±0.02 \\
~ & Precision & 0.600±0.07 & 0.621±0.06 & 0.749±0.05 & 0.768±0.04 & 0.746±0.03 & 0.716±0.02 & 0.796±0.01 & 0.791±0.00 \\
~ & Recall    & 0.575±0.09 & 0.583±0.08 & 0.570±0.07 & 0.580±0.06 & 0.781±0.05 & 0.739±0.04 & 0.715±0.03 & 0.768±0.02 \\
~ & F1        & 0.589±0.10 & 0.591±0.09 & 0.715±0.08 & 0.609±0.07 & 0.765±0.06 & 0.739±0.05 & 0.715±0.04 & 0.768±0.03 \\
\midrule
\multirow{4}{*}{\textbf{MTDAN}} 
& ACC       & 0.598±0.07 & 0.607±0.06 & 0.612±0.05 & 0.623±0.04 & 0.758±0.03 & 0.776±0.02 & 0.704±0.01 & 0.812±0.00 \\
~ & Precision & 0.592±0.08 & 0.614±0.07 & 0.618±0.06 & 0.637±0.05 & 0.796±0.04 & 0.803±0.03 & 0.729±0.02 & 0.825±0.01 \\
~ & Recall    & 0.573±0.11 & 0.608±0.09 & 0.598±0.08 & 0.582±0.07 & 0.772±0.06 & 0.797±0.05 & 0.731±0.04 & 0.796±0.03 \\
~ & F1        & 0.585±0.10 & 0.603±0.08 & 0.592±0.07 & 0.586±0.06 & 0.778±0.05 & 0.796±0.04 & 0.729±0.03 & 0.804±0.02 \\
\bottomrule
\end{tabular}
\end{table*}

To further validate that our SDI method does not merely remove difficult-to-classify samples but rather reduces conceptual ambiguity to facilitate more meaningful feature learning, we conducted an additional experiment. We set the SDI threshold to 0.1, using 90\% of the filtered data (conceptually clearer samples) as the training set and the remaining 10\% (samples filtered out due to higher discrepancy) as the testing set. Both sets were down-sampled to balance class distributions, and a RF model was employed. The results, presented in Table \textbf{\ref{train_test_sdi}}, demonstrate that the model achieves high performance even on these filtered-out samples. Accuracy, precision, recall, and F1 scores all exceeded 0.7 for Depression, Anxiety, and Stress factors.

This crucial finding indicates that the excluded samples still contain valuable and classifiable information; they are not simply invalid or noisy data points. Instead, these samples may represent novel diagnostic or risk categories---such as subthreshold Depression or high-risk populations---that warrant further investigation. Thus, SDI’s selective removal of samples with highly inconsistent labels enhances the coherence of the dataset for learning disorder-specific features. This, in turn, improves classification accuracy on the refined dataset while simultaneously highlighting the potential for data-driven approaches to support diagnostic innovation in affective disorders \cite{insel2010research, kotov2017hierarchical}.

In summary, both BC and RBC tasks initially suffered from the inherent overlap and variability in symptom severity within the complete dataset. The application of GDI-based filtering, guided by our SDI approach, substantially mitigated these issues. The RBC task, by design, further clarified class boundaries, leading to superior performance gains. The validation experiment confirmed that the SDI approach effectively isolates conceptually ambiguous samples rather than merely discarding difficult ones, pointing towards their potential clinical significance. This refined understanding and handling of data heterogeneity are critical for improving the generalization performance in emotion detection and affective disorder classification.

\begin{table}[!ht]
\centering
\renewcommand{\arraystretch}{1}
\caption{Excluded Samples Validate SDI-based Classification Performance.}
\label{train_test_sdi}
\setlength{\tabcolsep}{3pt}
\begin{tabular}{lccccccc}
\toprule
\textbf{Factor} & \textbf{TrainSize} & \textbf{TestSize} & \textbf{Accuracy} & \textbf{Precision} & \textbf{Recall} & \textbf{F1} \\
\midrule
Depression & 7307 & 823 & 0.733 & 0.760 & 0.680 & 0.718 \\
Anxiety & 7304 & 826 & 0.719 & 0.734 & 0.687 & 0.710 \\
Stress & 7305 & 825 & 0.717 & 0.732 & 0.684 & 0.708 \\
\bottomrule
\end{tabular}
\end{table}

\subsection{Regression Performance}
In the regression analysis, we analyzed the efficacy of these four models, across three psychological factors: Depression, Anxiety, and Stress. We also set different thresholds, like the GDI-based filter in classification tasks. The results are shown in \textbf{Table \ref{table_regression_all}}.

The implementation of our proposed GDI-based filtering method led to substantial improvements in model performance, significantly reduced prediction errors. For instance, applying a 0.1 threshold to the SVM model with statistical features decreased the MAE for Depression from 4.961 to 3.024, a remarkable 39\% reduction.

The image frame-based MSN approach, combined with our GDI filtering method, demonstrated superior performance. At the 0.1 threshold, it achieved the lowest MAEs of 2.701 and 3.228 for Depression and Stress. This consistently outperformed other models, including LI-FPN and MTDAN. The MTDAN has a lower MAE of 2.698 at the 0.1 threshold for Anxiety. 

\begin{table}[htbp]
    \caption{Regression Results of Multi-Model Evaluation With Thresholds}
    \renewcommand{\arraystretch}{1}
    \label{table_regression_all}
    \setlength{\tabcolsep}{3pt}
    \centering
    \footnotesize
    \begin{tabular}{c | c c | c c | c c | c c}
        \toprule
        \multirow{2}{*}{\textbf{Feature}} & \multirow{2}{*}{\textbf{Algorithm}} & \multirow{2}{*}{\textbf{Metric}} & \multicolumn{2}{c|}{\textbf{Depression}} & \multicolumn{2}{c|}{\textbf{Anxiety}} & \multicolumn{2}{c}{\textbf{Stress}} \\
        \cmidrule(lr){4-9}
        & & & 0 & 0.1 & 0 & 0.1 & 0 & 0.1 \\
        \midrule
        \multirow{4}{*}{Statistical} & \multirow{2}{*}{SVM} & MAE & 4.961 & 3.024 & 5.004 & 2.977 & 5.867 & 3.523 \\
        & & RMSE & 7.719 & 4.879 & 7.246 & 4.371 & 7.748 & 4.691 \\
        \cmidrule(lr){2-9}
        & \multirow{2}{*}{RF} & MAE & 5.697 & 2.888 & 5.550 & 3.017 & 6.155 & 3.736 \\
        & & RMSE & 7.546 & 3.914 & 7.117 & 3.912 & 7.679 & 4.68 \\
        \midrule
        \multirow{4}{*}{Image} & \multirow{2}{*}{LI-FPN} & MAE & 5.401 & 2.785 & 5.249 & 2.816 & 6.188 & 3.312 \\
        & & RMSE & 7.334 & 3.826 & 7.077 & 3.752 & 8.156 & 4.387 \\
        \cmidrule(lr){2-9}
        & \multirow{2}{*}{MSN} & MAE & 5.494 & 2.701 & 5.668 & 2.732 & 6.489 & 3.228 \\
        & & RMSE & 6.769 & 3.712 & 6.870 & 3.638 & 7.841 & 4.273 \\
        \cmidrule(lr){2-9}
        & \multirow{2}{*}{MTDAN} & MAE & 5.387 & 2.743 & 5.592 & 2.698 & 6.523 & 3.301 \\
        & & RMSE & 6.652 & 3.789 & 6.795 & 3.601 & 7.926 & 4.356 \\
        \toprule
    \end{tabular}
\end{table}

\subsection{Outlier Methods Comparison}

Our SDI method leverages heterogeneity by integrating multiple data sources and feature representations, which enhances its ability to identify complex outliers that single-method approaches might miss. In the conducted outlier methods comparison experiments on large-scale datasets such as FACES, we compared our method with several classical outlier detection methods including Isolation Forest \cite{liu2008isolation}, Local Outlier Factor (LOF) \cite{breunig2000lof}, and One-class SVM \cite{scholkopf1999support}. We set the contamination rate to approximately 10\% to maintain consistency across methods. After outlier removal, a RF model was used as the baseline for classification/regression tasks, the classification task followed the RBC experimental setup.

The results, presented in \textbf{Table \ref{outlier_comparison}}, indicate that all baseline methods improved performance compared to no outlier removal. Our SDI method demonstrates exceptional stability in both classification and regression tasks, achieving performance comparable to widely used outlier detection algorithms. Notably, IsolationForest is capable of providing consistent improvements similar to SDI. However, SDI stands out with its superior interpretability, particularly in the context of scale-based evaluations.

\begin{table*}[!ht]
\centering
\caption{Performance Comparison of Outlier Detection Methods.}
\renewcommand{\arraystretch}{1.1}
\label{outlier_comparison}
\begin{tabular}{lc|cccc|cc}
\toprule
\textbf{Outlier Method} & \textbf{Factor} & \textbf{Accuracy} & \textbf{Precision} & \textbf{Recall} & \textbf{F1 Score} & \textbf{MAE} & \textbf{RMSE} \\
\midrule
IsolationForest & Depression & 0.72 & 0.71 & 0.73 & 0.72 & 3.102 & 4.799 \\
IsolationForest & Anxiety     & 0.68 & 0.69 & 0.67 & 0.68 & 3.221 & 5.123 \\
IsolationForest & Stress      & 0.70 & 0.72 & 0.69 & 0.70 & 3.510 & 4.922 \\
\midrule
One-Class SVM    & Depression & 0.69 & 0.70 & 0.68 & 0.69 & 3.801 & 5.416 \\
One-Class SVM    & Anxiety     & 0.71 & 0.70 & 0.72 & 0.71 & 3.815 & 5.512 \\
One-Class SVM    & Stress      & 0.70 & 0.69 & 0.71 & 0.70 & 3.735 & 5.467 \\
\midrule
LOF & Depression & 0.73 & 0.74 & 0.72 & 0.73 & 3.909 & 5.678 \\
LOF & Anxiety     & 0.72 & 0.73 & 0.71 & 0.72 & 3.953 & 5.712 \\
LOF & Stress      & 0.74 & 0.75 & 0.73 & 0.74 & 4.000 & 5.890 \\
\midrule
\textbf{SDI (ours)} & \textbf{Depression} & \textbf{0.75} & \textbf{0.76} & \textbf{0.74} & \textbf{0.75} & \textbf{3.024} & \textbf{4.879} \\
\textbf{SDI (ours)} & \textbf{Anxiety}     & \textbf{0.73} & \textbf{0.73} & \textbf{0.79} & \textbf{0.76} & \textbf{2.977} & \textbf{4.371} \\
\textbf{SDI (ours)} & \textbf{Stress}      & \textbf{0.69} & \textbf{0.75} & \textbf{0.70} & \textbf{0.73} & \textbf{3.523} & \textbf{4.691} \\
\bottomrule
\end{tabular}
\end{table*}

\subsection{Demographic Difference Experiment}
To address potential demographic variations and ensure the robustness of our findings, we conducted an additional experiment examining gender differences in classification performance using the FACES dataset. The dataset was stratified by gender, creating male and female subsets with sample sizes of 4367 and 3914, respectively, and we evaluated the performance of our SDI and classification model for Depression, Anxiety, and Stress using accuracy, precision, recall, and F1 score. As presented in \textbf{Table \ref{table_gender}}, the results show negligible differences between males and females across all three factors: depression and anxiety metrics were nearly identical, while stress scores were slightly higher for females, though the variations were minimal. These findings indicate that the SDI and classification framework perform consistently across genders, supporting the method’s fairness and applicability in mental health assessments.

\begin{table}[!h]
\caption{Gender Classification Performance}
\footnotesize
\renewcommand{\arraystretch}{1.1}
\label{table_gender}
\centering
\begin{tabular}{lcccccc}
\toprule
Gender & Count & Factor & Accuracy & Precision & Recall & F1\\ \hline
\multirow{3}{*}{Male}   & \multirow{3}{*}{4367} & Depression & 0.788    & 0.803     & 0.788  & 0.793 \\
 & & Anxiety & 0.778 & 0.803 & 0.778  & 0.785 \\
 & & Stress & 0.668 & 0.752 & 0.668  & 0.696 \\
\midrule
\multirow{3}{*}{Female} & \multirow{3}{*}{3914} & Depression & 0.788 & 0.794 & 0.788 & 0.791 \\
 & & Anxiety & 0.793 & 0.803 & 0.793  & 0.796 \\
 & & Stress & 0.699 & 0.731 & 0.699 & 0.712\\
\bottomrule
\end{tabular}
\end{table}

\section{Discussion}

Adolescence is a distinct period marked by profound psychological changes, during which one in seven individuals will suffer from a mood disorder \cite{WHO_adolescent_mental_health}. Prompt and efficient screening for mood disorders is a crucial area of research focus. The present work set out to resolve two long standing obstacles in automated mood disorder assessment: the shortage of population scale facial corpora and the hidden heterogeneity that arises when multi-item scales are collapsed into single severity labels. By releasing the FACES data set and formulating the Symptom Discrepancy Index, we were able to quantify and then reduce the label noise that has hindered previous efforts \cite{wang_fast_2023,xu_attention-based_2023}.

Our results show that computer vision models trained on reading videos can reach F1 values near 0.80 for binary psychopathology detection once symptom heterogeneity is properly controlled. These results demonstrate the reliability comparable to traditional questionnaires, thereby supporting their use as a practical and objective approach to cost-effective school-based screening. Across five algorithmic families—from random forests to multi-task spatiotemporal networks—selectively removing the 10\% of records with the highest SDI scores raised F1 by 25–30\% and lowered regression MAE by about 40\%. Because identical gains were observed on the retained and on the discarded partitions, SDI acts by clarifying the mapping between facial cues and symptom patterns rather than by censoring “hard’’ examples. This finding revises the widespread assumption that poor generalization on large, naturalistic corpora is primarily an architectural problem; much of the deficit can instead be traced to psychometric label construction.

SDI builds on recent network psychology theories \cite{pan2024module} that conceptualize disorders as groups of interacting symptoms. Whereas those theories focus on estimating edges in a symptom graph, SDI offers a single numerical value that can be computed rapidly and compared across samples, scales and time points. The normalized 0–1 range facilitates intuitive interpretation: low values indicate homogeneous symptom patterns, whereas high values suggest highly varied profiles that will likely interfere with classification accuracy.

Several important limitations should be noted. First, FACES currently represents a single cultural context (East Asian adolescents). Cross cultural replications are necessary to establish ecological validity. Second, the linear distance metric used in SDI may miss non linear symptom interactions; kernelized or manifold based variants are an obvious extension. Third, GDI estimation can become unstable when very few pupils share the same total score; Bayesian adjustment or neighborhood pooling may alleviate this issue.

Three future directions appear especially promising. First, longitudinal modeling that exploits the repeated measurement design of FACES to predict future symptom trajectories; Second, multi modal fusion of facial, vocal and textual signals available in the cohort; Third, integration of SDI into active learning pipelines that request additional labels precisely for the most heterogeneous cases, thereby maximizing annotation efficiency.

\section{Conclusion}

This study contributes a two part advance to affective computing and adolescent mental health screening. First, we release FACES, the largest and most standardized video corpus of its kind, offering sufficient statistical power for modern deep architectures and for rigorous out of distribution testing. Second, we formalize the Symptom Discrepancy Index, a simple yet powerful measure that exposes the item level variability hidden by traditional total score labels. Experiments demonstrate that SDI guided filtering alone can boost diagnostic F1 from about 0.50 to 0.80 and halve regression error, without altering model architectures. These findings establish symptom heterogeneity—not model capacity—as the principal bottleneck in large scale automated assessment and provide a general remedy that is readily applicable to other psychometric data sets. By combining a population level resource with a principled heterogeneity metric, the work lays the groundwork for precise, scalable and ethically responsible mental health monitoring in real world settings.

\section{Data Availability}

The FACES were collected with full ethical compliance and approval from the Medical Research Ethics Committee of the Affiliated Brain Hospital of Nanjing Medical University (2022-KY095-02). All participants provided written informed consent, ensuring their understanding of the study’s purpose and data usage. The updates on data accessibility can be found at \url{https://github.com/njnklab/FACES}.

\bibliographystyle{IEEEtran}
\bibliography{References}{}

\vspace{-1cm}
\begin{IEEEbiography}[{\includegraphics[width=1in,height=1.25in,clip,keepaspectratio]{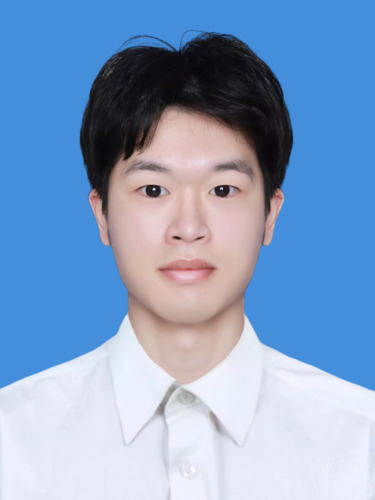}}]{Xiao Xu}
received his B.Eng degree in Software Engineering at Nanjing University of Chinese Medicine and his M.Eng degree in Biomedical Engineering from Nanjing Medical University (NJMU), China. He is currently pursuing a BME Ph.D. degree at NJMU. His main research interests lie at the intersection of audio-visual technology, artificial intelligence in medicine.
\end{IEEEbiography}
\vspace{-1cm}
\begin{IEEEbiography}[{\includegraphics[width=1in,height=1.25in,clip,keepaspectratio]{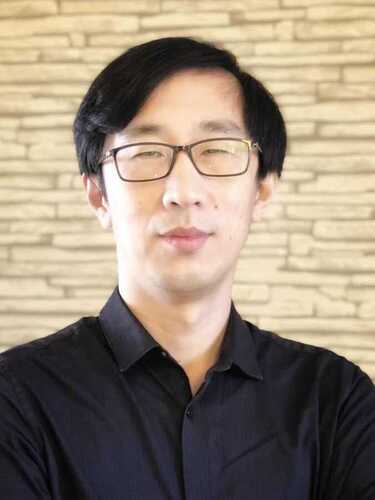}}]{Xizhe Zhang}
received his Ph.D. in Computer Science and Technology from Jilin University, China, in 2006. That same year, he secured a position as an Associate Professor at School of Computer Science and Engineering of Northeastern University. In 2020, he became a Full Professor at Nanjing Medical University and Nanjing Brain Hospital, concentrating on the interdisciplinary field of artificial intelligence and brain science. He has received prestigious recognitions, including being named a Jiangsu Special Medical Expert and Distinguished Membership in the China Computer Federation. His research endeavors are primarily focused on machine learning, artificial intelligence, complex network analysis, and their medical applications.
\end{IEEEbiography}
\vspace{-2cm}
\begin{IEEEbiography}[{\includegraphics[width=1in,height=1.25in,clip,keepaspectratio]{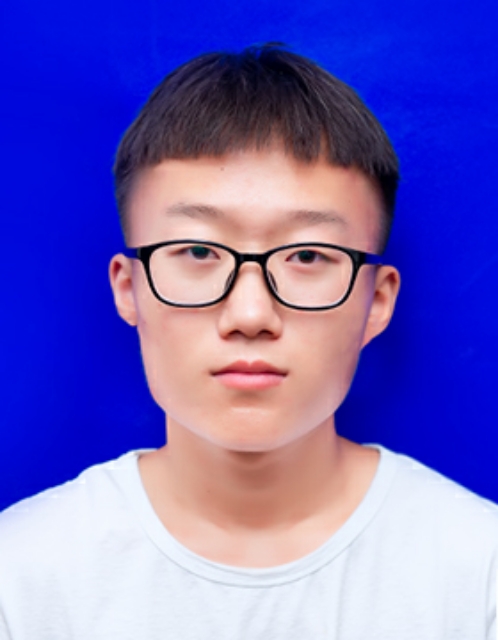}}]{Yan Zhang}
received his Bachelor's degree in Medical Information Engineering from Nanjing University of Chinese Medicine in China in 2019. He is currently pursuing a master's degree in biomedical engineering at Nanjing Medical University in China.  His research interests focus on the application of brain image preprocessing and analysis workflow to the study of mental disorders.
\end{IEEEbiography}

\end{document}